\documentclass[acmtog]{acmart}
\AtBeginDocument{%
  }

\copyrightyear{2025}
\acmYear{2025}
\setcopyright{acmlicensed}\acmConference[SA Conference Papers '25]{SIGGRAPH Asia 2025 Conference Papers}{December 15--18, 2025}{Hong Kong, Hong Kong}
\acmBooktitle{SIGGRAPH Asia 2025 Conference Papers (SA Conference Papers '25), December 15--18, 2025, Hong Kong, Hong Kong}
\acmDOI{10.1145/3757377.3763855}
\acmISBN{979-8-4007-2137-3/2025/12}

\usepackage{multirow}
\usepackage{pifont}
\usepackage{bbding}
\usepackage{makecell}

\newcommand{\etal}{{\emph{et al.}}}
\newcommand{\datasetname}{{\sc SKEP-120K }}

\acmSubmissionID{papers\_1307}

\begin{document}

\title{Sketch2PoseNet: Efficient and Generalized Sketch to 3D Human Pose Prediction}



\author{Li Wang}
\affiliation{%
  \institution{Nanjing University}
  \city{Nanjing}
  \country{China}}
\email{liwang1029@smail.nju.edu.cn}
\orcid{0009-0002-0186-6028}

\author{Yiyu Zhuang}
\affiliation{%
  \institution{Nanjing University}
  \city{Nanjing}
  \country{China}}
\orcid{0009-0003-5726-8534}

\author{Yanwen Wang}
\affiliation{%
  \institution{Nanjing University}
  \city{Nanjing}
  \country{China}}
\orcid{0009-0000-7361-1679}

\author{Xun Cao}
\affiliation{%
  \institution{Nanjing University}
  \city{Nanjing}
  \country{China}}
\orcid{0000-0003-3094-4371}

\author{Chuan Guo}
\affiliation{%
  \institution{Snap Inc.}
  \city{New York}
  \country{USA}}
\orcid{0000-0002-4539-0634}

\author{Xinxin Zuo}
\affiliation{%
  \institution{Concordia University}
  \city{Montreal}
  \country{Canada}}
\orcid{0000-0002-7116-9634}

\author{Hao Zhu}
\affiliation{%
  \institution{Nanjing University}
  \city{Nanjing}
  \country{China}}
\orcid{0000-0003-1596-4366}
\authornote{Corresponding Author.}

\renewcommand{\shortauthors}{Wang et al.}




\begin{teaserfigure}
  \includegraphics[width=\textwidth]{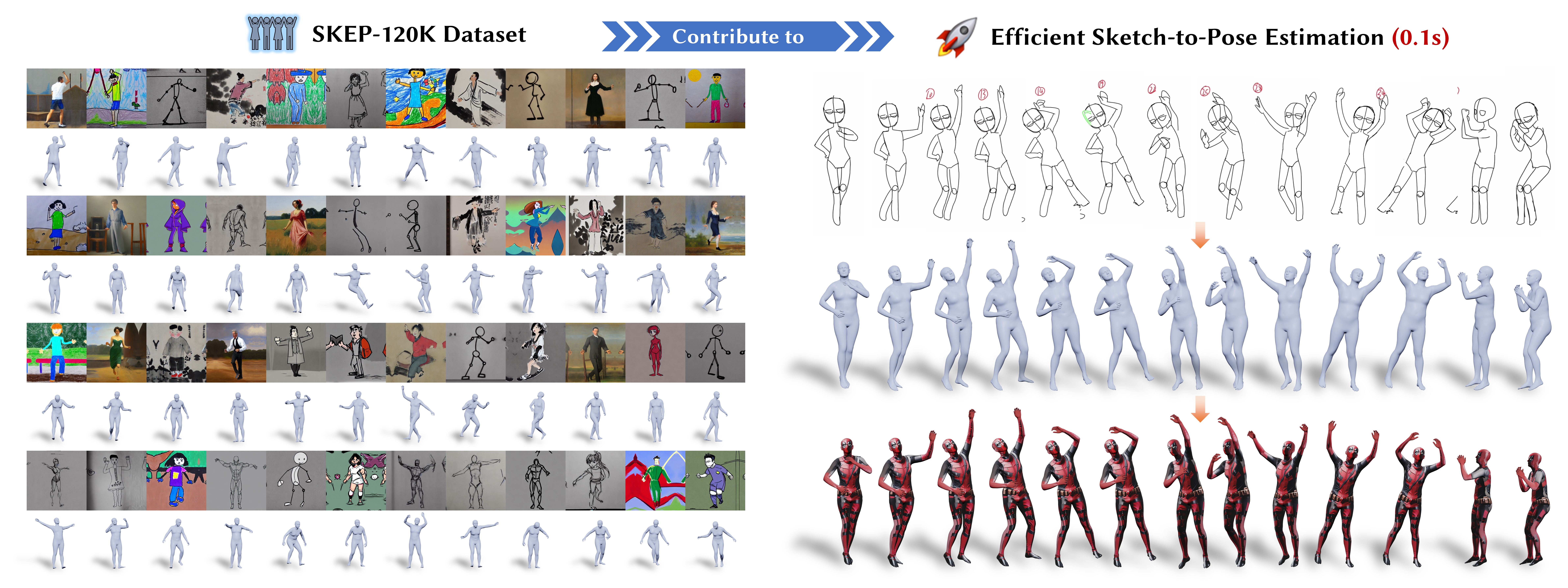}
  \caption{We present a novel approach for 3D human pose estimation from sketches. Benefiting from the large-scale \datasetname dataset (left), we propose to learn a data-driven sketch-to-pose model that exhibits improved generalization ability and efficient inference (right).}
  \label{fig:teaser}
\end{teaserfigure}


\begin{abstract}
3D human pose estimation from sketches has broad applications in computer animation and film production. Unlike traditional human pose estimation, this task presents unique challenges due to the abstract and disproportionate nature of sketches. Previous sketch-to-pose methods, constrained by the lack of large-scale sketch-3D pose annotations, primarily relied on optimization with heuristic rules—an approach that is both time-consuming and limited in generalizability. To address these challenges, we propose a novel approach leveraging a "\textit{learn from synthesis}" strategy. 
Firstly, a diffusion model is learned to synthesize sketch images from 2D poses projected from 3D human poses, mimicking disproportionate human structures in sketches.
This process enables the creation of a synthetic dataset, \datasetname, consisting of \textit{120k} accurate sketch-3D pose annotation pairs across various sketch styles.
Building on this synthetic dataset, we introduce an end-to-end data-driven framework for estimating human poses and shapes from diverse sketch styles. Our framework combines existing 2D pose detectors and generative diffusion priors for sketch feature extraction with a feed-forward neural network for efficient 2D pose estimation. 
Multiple heuristic loss functions have been incorporated to guarantee geometric coherence between the derived 3D poses and the detected 2D poses while preserving accurate self-contacts.
Qualitative, quantitative, and subjective evaluations collectively affirm that our proposed model substantially surpasses previous ones in both estimation accuracy and speed for sketch-to-pose tasks.
\end{abstract}

\maketitle

\section{Introduction}
\label{sec:intro}

Human pose estimation holds significant importance and finds widespread application across numerous scenarios, including 3D human reconstruction~\cite{tan2020self, zhu2016video}, 3D human generation~\cite{zhuang2025idol, wang2025tera, zeng2023avatarbooth}, view synthesis~\cite{zhu2018view}, and animation~\cite{zhu2024champ, liao2020speech2video}.  Among the various sources used for pose estimation, sketches emerge as an efficient and versatile entity. Sketches are data that can be more easily designed by artists and are widely used in animation and film production.  More broadly, the term `sketch' encompasses a diverse range of graphical styles, including charcoal sketches, cartoons, stick figures, kids' drawings, oil paintings, ink paintings, and so forth.

Estimating human poses from sketches presents a significant challenge. Generalized photo-based pose estimation methods fall short in this task due to their exclusive training on realistic data.    By contrast, sketches often disregard human proportionality and geometric perspective, opting for a more abstract representation of poses, thereby exacerbating the complexity of the sketch-to-pose conversion.  To tackle this, Brodt \etal introduced Sketch2Pose \cite{brodt2022sketch2pose}, which initializes by predicting 2D joint positions from sketches and subsequently aligns a 3D parametric human model to their bones via an optimization framework.  Nonetheless, this method is slow and mostly tailored towards hand-drawn sketch lines.  Pursuing a swift and highly generalized solution for the sketch-to-pose task remains an open problem.

To tackle this problem, we embraced a "learn from synthesis" strategy, which has been successfully applied in avatar modeling~\cite{guo2023rafare, zhuang2024towards} and street view synthesis~\cite{zhu2024streetsyn}. Starting from a modest quantity of sketches and corresponding 2D human pose datasets, a large-scale sketch-3D pose dataset is synthesized by a fine-tuned image generative model conditioned on human poses. Such data synthesis is tailored for the sketch-to-pose task. Specifically, we incorporated pose perturbations to create data representing disproportionate human figures and misaligned perspectives in sketches. Furthermore, we amassed a substantial collection of sketches encompassing diverse styles, conducted detailed categorical analyses, and thereby enriched the stylistic variety of the sketches we generated. Ultimately, we produced $120,000$ such high-quality sketch-pose data pairs.

Based on such a dataset, we introduce an end-to-end framework for estimating human mesh from various styled sketches. The generative diffusion prior is leveraged to extract human pose features in sketches and inject conditions that fit the drawing features to guide the denoising network. Unlike the iterative optimization strategy utilized by Sketch2Pose, we implement a neural network featuring a feed-forward architecture for almost 500 times faster pose estimation.  A feature-extracting strategy tailored for sketches is introduced to boost the accuracy of 3D pose regression.  Owing to our extensive dataset encompassing a wide range of styles and a meticulously designed loss function, our method achieves comparable pose estimation accuracy to Sketch2Pose, while significantly surpassing it in terms of speed and generalization capabilities. 


The contributions of our work can be summarized as follows:
\begin{itemize}
    \item Using a learning-by-synthesizing strategy, we propose a novel approach to address the sketch-to-pose problem. This strategy involves synthesizing a large-scale, customized sketch-3D pose dataset, which substantially boosts the generalization capabilities of the sketch-to-pose estimator across diverse sketch styles.
    \item By developing a feed-forward structured network, we have significantly improved the speed of sketch-to-pose estimation, marking the 500 times faster than the prior SOTA sketch-to-pose estimator.
    \item Our meticulously designed network architecture and loss function have greatly enhanced the robustness of the prediction model, allowing it to accurately predict poses even in the presence of human proportion distortions and perspective inaccuracies that commonly exist in sketches. As a result, our method achieves state-of-the-art (SOTA) pose prediction accuracy.
\end{itemize}

\section{Related Works}
\label{sec:related}




Sketching is widely regarded as an easy and accessible way to pose characters, catering to both professionals and non-artists. 
While notable progress has been made in related fields, such as sketch-based interfaces and image-based pose estimations, the unique challenges of handling abstract, disproportionate, and stylistically diverse sketches remain underexplored. This section reviews the most relevant works, categorized into sketch-based character posing and human pose estimation from a single photograph.

\subsection{Sketch-Based Character Posing}
Sketch-based character posing provides an intuitive means for users to manipulate 3D human poses, yet it introduces several significant challenges.
Depth ambiguity, anatomical distortions, missing details, and diverse sketching styles make pose inference particularly difficult. 
Early works focused on stick figures~\cite{Hecker1992,Davis2003,Mao2005a,Lin2010}, silhouettes~\cite{shadowTheatre}, and clean vector drawings~\cite{Bessm2016:10.1145/2980179.2980240}. These approaches, though efficient in constrained scenarios, are hindered by their reliance on unambiguous, clean inputs. For instance, Gesture3D~\cite{Bessm2016:10.1145/2980179.2980240} reconstructs poses from vector drawings but assumes minimal noise, precise connectivity, and no extra strokes—requirements that are rarely met by natural, user-drawn sketches. This reliance on specific input types significantly limits the usability of such systems, as users cannot freely use diverse sketching styles to specify desired poses. 

Recent approaches like Sketch2Pose~\cite{brodt2022sketch2pose} use a neural network to predict bitmap representations and optimize 3D model parameters for pose inference. 
 However, due to the scarcity of sketch-to-3D model pairs for training, the method requires additional optimization to produce acceptable results. This not only introduces a significant computational burden, but also raises concerns about the reliability of the generated poses, which may lack naturalness or anatomical correctness.

An important application in this domain is the development of interactive systems for engaging with sketches. To achieve efficient inference, systems like MonsterMash~\cite{dvorovzvnak2020monster} and Motion Doodles~\cite{thorne2004motion} offer fast, intuitive sketch-based interactions but are limited by strict input formats or detailed annotations. Systems designed for articulated human poses, like those by Unlu \etal\cite{unlu2022interactive} and Schmitz \etal\cite{schmitz2023interactive}, impose further input constraints, requiring sketches to consist of 3D primitives. 
Previous methods have imposed a trade-off between efficiency and input diversity due to the lack of paired sketch-3D pose datasets. In contrast, our approach addresses this gap by proposing a large-scale dataset and building a pose estimation network that directly predicts poses from sketches. This enables efficient near-real-time performance while maintaining generalizability, offering a simple, direct, and scalable solution for sketch-to-pose estimation.

\begin{figure*}[!htbp]
    \centering
    \includegraphics[width=1.0\linewidth]{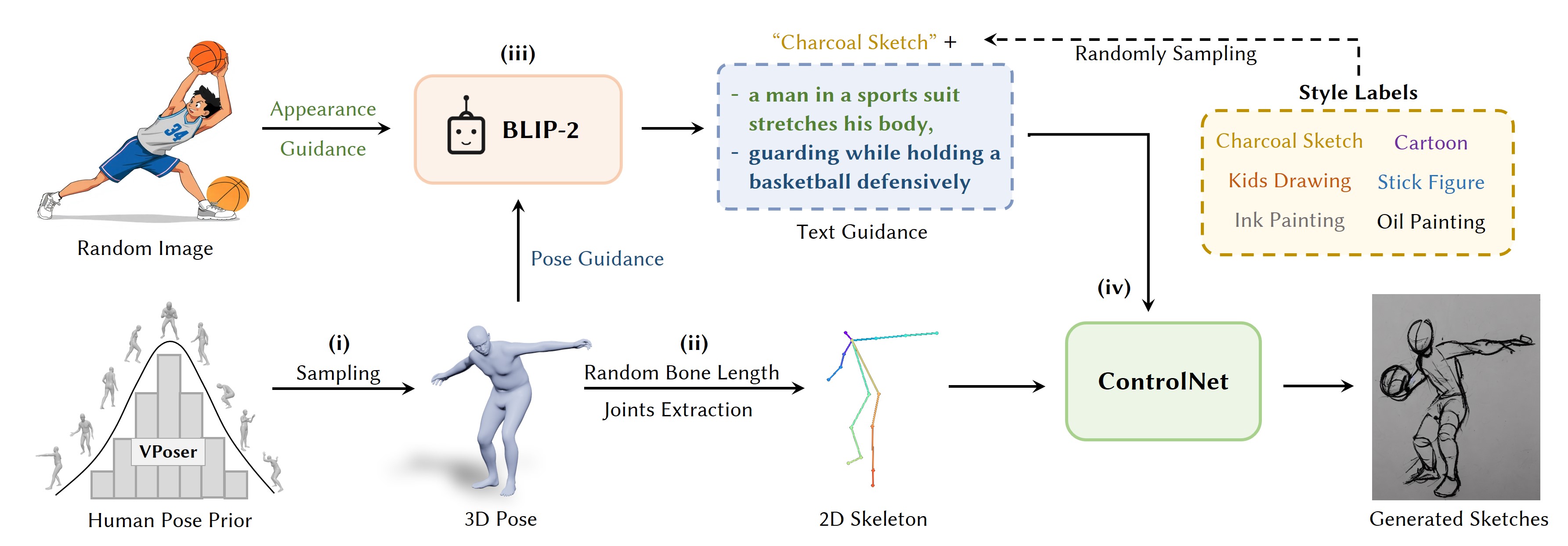}
    \caption{\datasetname Dataset Creating Pipeline. Three stages are involved: (I) generating diverse 3D poses (as SMPL); (II) adding random biases to bone lengths and projecting to 2D poses; (III) generating diverse text guidance; (IV) training a text-conditioned image generator for sketch synthesis.}
    \label{fig:data_gen}
\end{figure*}

\subsection{Human Pose from a Single Photograph}

Estimating 3D human poses from a monocular image has been extensively studied in computer vision due to its significant applications in computer graphics, animation, and human-computer interaction. 
Early methods relied on handcrafted features~\cite{balan2007detailed,Ramanan2011,andriluka2010monocular}, using probabilistic models and tree-based structures. However, they struggled with occlusions, ambiguous poses, and appearance variations.

The introduction of deep learning shifted the field, with DeepPose~\cite{toshev2014deeppose} being one of the first CNN-based approaches. This was followed by methods like Tekin \etal~\cite{Tekin_2016_CVPR}, which integrated CNNs with structured prediction to improve pose accuracy, and Martinez \etal~\cite{martinez_2017_3dbaseline}, which proposed a fully connected network for 2D-to-3D lifting. Zhou \etal~\cite{Zhou_2017_ICCV} added geometric constraints, while Pavlakos \etal~\cite{pavlakos2017coarse} used volumetric heatmaps for joint localization.



Despite progress, the need for large labeled datasets limited generalization, especially for non-photorealistic inputs. The introduction of parametric models like SMPL~\cite{bogo2016keep} and SMPL-X~\cite{pavlakos2019expressive} advanced pose and shape estimation with 3D human priors. Many methods~\cite{zhang2021pymaf, li2021hybrik, li2022cliff, goel2023humans} focus on improving the accuracy of human mesh recovery. 
Weakly supervised approaches like HMR~\cite{KanazawaBJM18} regressed SMPL parameters using 2D keypoints and adversarial losses, and HMD~\cite{zhu2021detailed, zhu2019detailed} further refined detailed shape based on the predicted SMPL mesh. Kolotouros~\etal's SPIN~\cite{kolotouros2019spin} refined this approach with optimization, while EFT~\cite{joo2020eft} fine-tuned SMPL predictions. Methods like 3DCrowdNet~\cite{choi2022learning}, JOTR~\cite{li2023jotr}, and DPMesh~\cite{zhu2024dpmesh} are designed to recover the occluded body mesh.
Self-supervised methods, such as Wang \etal~\cite{wang20193d} and Novotny \etal~\cite{Novotny2019}, reduced dependence on labeled data using geometric consistency.


Our method bridges human pose estimation and character drawing by leveraging visual priors in pre-trained diffusion models, as seen in VPD~\cite{zhao2023unleashing}. Using our dataset, our fine-tuned network extracts structural and spatial features from the denoising U-Net for accurate human mesh recovery. By processing the input image in a single inference pass~\cite{zhu2024dpmesh}, our approach adapts diffusion priors to handle abstract sketches, ensuring reliable pose estimation.


\begin{figure}[!h]
    \centering
    \includegraphics[width=1.0\linewidth]{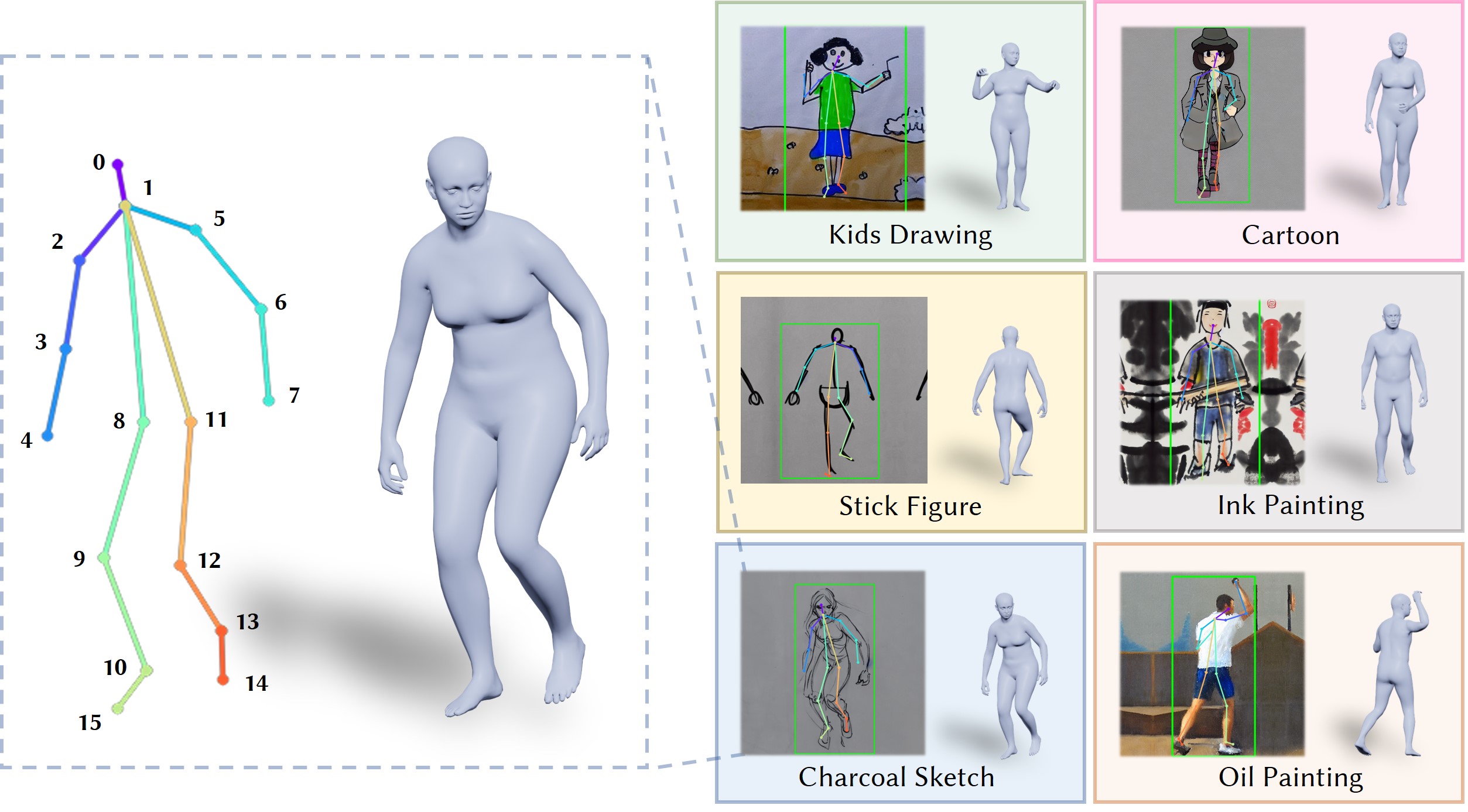}
    \caption{Data Description. \datasetname dataset comprises six sketch styles: cartoons, oil paintings, ink paintings, charcoal sketches, stick figures, and kids' drawings. The provided 2D/3D joints are shown on the left.
    }
\label{fig:data_desc}
\end{figure}
\section{\datasetname Dataset}\label{sec: dataset}



It is widely recognized that the quality and abundance of the training data heavily influence the success of learning-based techniques for human mesh recovery.  Surprisingly, we find that there is currently a notable absence of a large-scale, high-quality dataset containing sketches and 3D human poses. Though several available datasets~\cite{madhu2022enhancing, brodt2022sketch2pose, smith2023method, ju2023human} offer a substantial number of sketches, they provide only 2D pose labels and typically a single sketch style, making them insufficient for training a highly accurate, generalizable sketch-to-pose model.


Therefore, we propose a Sketch and 3D Pose dataset with $120k$ data pairs in various sketch styles, named as \datasetname dataset.  
As shown in Fig.~\ref{fig:data_desc}, the dataset encompasses six styles according to artificial human scenes: \textit{cartoons, oil paintings, ink paintings, charcoal sketches, stick figures, and kids' drawings}. Each style contains approximately 20,000 images. Our dataset provides human bounding boxes, 16 human joints (both 2D and 3D, with corresponding visible/invisible/included attributes), SMPL pose parameters, and text information.
Due to the different definitions of the skeleton for a human pose in different datasets and considering the characteristics of gesture expression in the sketch, we define a new 3D skeleton to represent the human pose.  Specifically, the body parts are based on MSCOCO database~\cite{lin2014microsoft}, and two additional joints regarding left and right toes are added to reflect fine-scale leg poses. 


The creation process of our dataset is shown in Fig.~\ref{fig:data_gen}. Firstly, we utilize VPoser~\cite{pavlakos2019expressive} to generate random SMPL models exhibiting diverse and plausible poses, where a variational autoencoder is designed to capture latent representations of human poses.  
{Given the pervasive foreshortening in sketches, which makes the depicted human structure to deviate from the standard 3D-to-2D projection observed in real-captured images, we introduce skeletal-proportion perturbations during the 3D-to-2D mapping by adding random biases to the projected limb lengths as a skeleton-level data augmentation. This strategy yields skeletal configurations that better capture the characteristic proportional exaggerations and imbalances of hand-drawn sketches, thereby enhancing the pose diversity and accuracy of the sketch dataset and producing 2D joint annotations that are better aligned with sketch scenarios.}
{Next, we consolidate data from the Sketch2Pose ~\cite{brodt2022sketch2pose}, Human-Art ~\cite{ju2023human}, and Amateur Drawing datasets ~\cite{smith2023method}, unify their 2D keypoint annotations to our defined set of 16 joints, and partition them into six groups by drawing style. We then leverage BLIP2~\cite{li2023blip} to generate appearance descriptions on sketch images and motion descriptions on SMPL rendering images. These descriptions, together with the target style labels, serve as conditioning prompts, yielding a sketch training set. On this basis, we train a text-conditioned image generation model following ControlNet~\cite{zhang2023adding} to synthesize sketch data. During training, sketches from all six styles and their associated annotation parameters are jointly used to train the same model.}

{After that, we manually curate the model-generated dataset to improve its quality further. Specifically, we invite experienced 3D modelers to filter out approximately $10\%$ of the sketches in each style, including those with severely cluttered backgrounds that compromise subject visibility, those showing pose inconsistencies caused by overly challenging conditioning poses, and a subset with extremely distorted poses introduced by the added random biases.} As a result, we generate character images across various styles with high accuracy and adherence to the 2D skeleton distribution. 
Given the diverse line-based nature of sketches, we use the off-the-shelf outline detector~\cite{canny1986computational} to extract line distributions within these images. A threshold is applied to identify the smallest region encompassing most lines, defining the character’s bounding box. We compare the human bounding box with the bounding box generated from the detector of Human-Art, leaving a more accurate result, and manually filtering out the undetectable cases.  We also detect the occlusion of each joint based on the occlusion relationship of the SMPL mesh, which is recorded as the label of each joint.
\section{Method}
\label{sec:method}

\begin{figure*}
    \centering
    \includegraphics[width=1.0\linewidth]{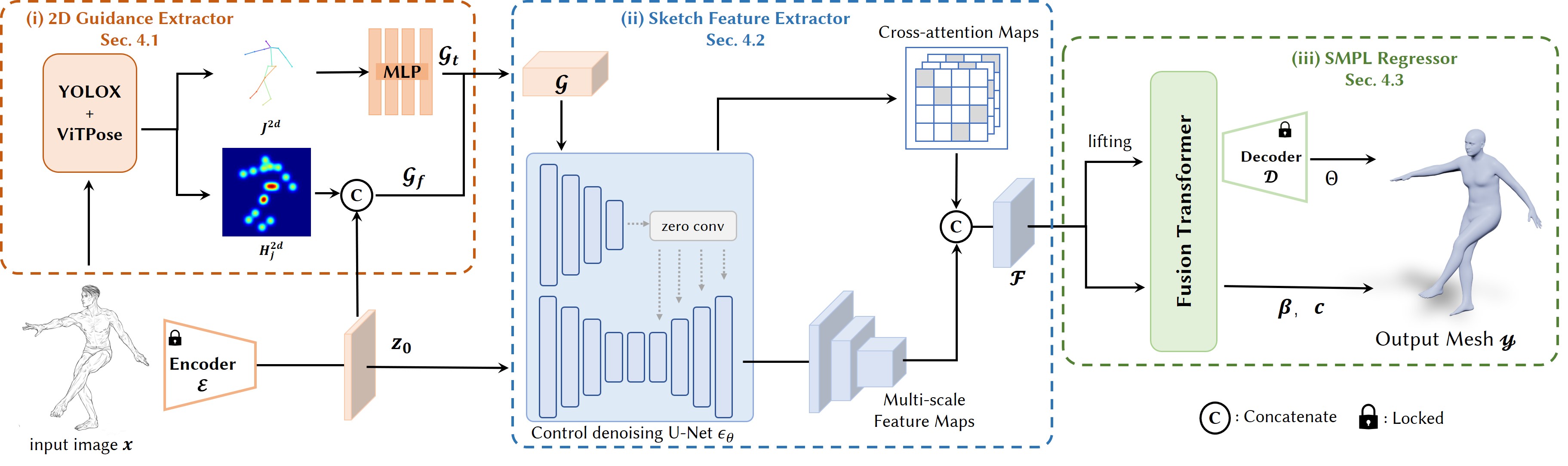}
    \caption{Overall Pipeline.  Given a sketch image as input, the network predicts 3D human poses represented by SMPL parameters.  The overall network consists of three modules: a 2D guidance extractor as detailed in Sec.~\ref{sec: guidance}; a sketch feature extractor as detailed in Sec.~\ref{sec: feature}; and an SMPL regressor as detailed in Sec.~\ref{sec: regressor}.}

    \label{fig:pipeline}
\end{figure*}

Given the \datasetname dataset, our objective is to train a prediction model for recovering 3D human poses from sketches in varying styles.  As shown in Fig.~\ref{fig:pipeline}, our overall network consists of three modules: (I) a 2D guidance extractor (Sec.~\ref{sec: guidance}); a sketch feature extractor (Sec.~\ref{sec: feature}); and an SMPL regressor (Sec.~\ref{sec: regressor}).  From a probabilistic model perspective, the above process can be formulated as:
\begin{equation}
    {p}_\phi(\boldsymbol{y} \!\mid\! \boldsymbol{x}) = p_{\phi_3}(\boldsymbol{y} \!\mid\! \mathcal{F}) p_{\phi_2}(\mathcal{F} \!\mid\! \epsilon(\boldsymbol{x}), \mathcal{G}) p_{\phi_1}(\mathcal{G} \!\mid\! \epsilon(\boldsymbol{x})),\label{eq:prob}
\end{equation}
where $x$ denotes the input sketch; $y$ is the 3D pose represented by SMPL parameters; $\mathcal{F}$ signifies informative feature maps extracted from sketches; $\mathcal{G}$ indicates the spatial guidance extracted from 2D poses; $\epsilon$ is a pre-trained image encoder network; $p_{\phi_1}$, $p_{\phi_2}$ and $p_{\phi_3}$ correspond to the 2D guidance extractor, sketch feature extractor, and SMPL regressor, respectively.  We will then explain each module and the objective functions in the following sections.

\subsection{2D Guidance Extractor}\label{sec: guidance}

Drawing inspiration from VPD~\cite{zhao2023unleashing}, our core idea involves extracting high-level pre-trained knowledge from a diffusion model. A fundamental prerequisite for achieving this is the extraction of 2D guidance.




The first step in extracting 2D guidance is to estimate 2D joints from input sketches.  Leveraging our proposed \datasetname dataset, we have fine-tuned two state-of-the-art network models for human detection and 2D joint extraction from sketches. Specifically, the input sketches are first resized and padded to the resolution of $256\times192$ pixels to preserve the aspect ratio. Then, the human detector YOLOX~\cite{redmon2016you} is fine-tuned by Human-Art dataset for human's bounding box detection in sketches, and the ViTPose~\cite{xu2022vitpose} model is fine-tuned for 2D joints prediction from these bounded sketches. ViTPose utilizes a straightforward, non-hierarchical vision transformer as the encoder to capture human features in drawings, combined with a lightweight decoder that predicts body joints in a top-down approach. Finally, we obtain 2D joints $J^{2D}\in \mathbb{R}^{K \times 2}$ along with their corresponding confidence and transform them into heatmaps $ H_j^{2D} \in \mathbb{R}^{K \times H^{'} \times W^{'}}$ using 2D Gaussian kernels~\cite{cheng2020higherhrnet}.

After the 2D joints are obtained, pose features are extracted from the 2D joints $J^{2D}$ and heatmaps $H_j^{2D}$, which provides spatial guidance for the denoising U-Net~\cite{rombach2022high} backbone $\boldsymbol{\epsilon_\theta}$. This process is referred as $p_{\phi_1}(\mathcal{G} \!\mid\! \boldsymbol{x})$ in Eq.~\ref{eq:prob}.  For the input image $\boldsymbol{x}$, we convert the cropped image $\boldsymbol{x} \in \mathbb{R}^{H \times W \times C}$ from pixel space to the latent space with frozen encoder $\mathcal{E}$ in the trained VQGAN from the Controlnet framework to obtain the latent representation $z_0 \in \mathbb{R}^{ H^{'} \times W^{'} \times \mathcal{G}_z}$. Then, we concatenate the heatmap $H_j^{2D}$ with the input image $z_0$ to obtain $ \mathcal{G}_f \in \mathbb{R}^{(K+{\mathcal{G}_z}) \times H^{'} \times W^{'}}$. In most previous diffusion models ~\cite{zhang2023adding, rombach2022high, dhariwal2021diffusion}, the prompt guidance $\mathcal{G}_t$  usually relies on text embeddings derived from a frozen CLIP~\cite{radford2021learning} model.  In contrast, we replace the text with 2D joint positions $J^{2D}$ as verified in ~\cite{zhu2024dpmesh}. To match the text token dimension $D_j$, a two-layer MLP is used to enhance the dimensionality of the 2D joint positions to 768 in the pre-trained diffusion model. This generates a spatial guidance $\mathcal{G}_t\in \mathbb{R}^{K \times D_j}$. The process can be expressed as follows:
\begin{equation}
    \mathcal{G}_f = \mathrm{Concat}({z}_0, \mathbf H_j^{2D}),
\end{equation}
\begin{equation}
    \mathcal{G}_t = \mathrm{MLP}({J}_{2D}),
\end{equation}
After that,  $\mathcal{G}_f$ and $\mathcal{G}_t$ are injected into $\boldsymbol{\epsilon_\theta}$ through different channels, thus we obtain the 2D guidance $\mathcal{G}$.

\subsection{Sketch Feature Extractor} \label{sec: feature}


Once the 2D guidance is obtained, our next objective is to extract informative features from the sketches for 3D pose estimation. A multi-scale features extractor is introduced based on the pre-trained denoising U-Net.
Our key idea is to fully extract the pre-trained high-level knowledge from a pretrained diffusion model, named informative features $\mathcal{F}$, then utilize its learned knowledge to predict 3D human poses from sketches. We employ the denoising U-Net $\boldsymbol{\epsilon_\theta}$ as the image backbone, performing a single inference to extract features from image $\boldsymbol{x}$.  To provide effective guidance, we utilize the conditional injection of human pose instead of the text condition, which makes the connection between these conditions and the input image such that the learned semantic information can be efficiently extracted. 




Specifically, $p_{\phi_2}(\mathcal{F} \!\mid\! \epsilon(\boldsymbol{x}), \mathcal{G})$ is designed to extract hierarchical feature maps $\mathcal{F}$ from the input image $\boldsymbol{x}$ along with the 2D guidance $\mathcal{G}$. We observe that the pre-trained text-to-image diffusion model serves as an excellent initialization for $p_{\phi_2}$, which has already established a connection between the vision and language domains. 

It is also known that ControlNet leverages trainable copies of the encoding layers within the denoising U-Net, serving as a robust backbone to learn various conditional controls, significantly enhancing the fine-grained spatial controllability of the Latent Diffusion Model (LDM) ~\cite{rombach2022high}.  
In our implementation, we utilize the ControlNet architecture to handle pose-conditioned information from the 2D guidance $\mathcal{G}$ and integrate it into the image features within the denoising U-Net $\boldsymbol{\epsilon_\theta}$. The output $\mathcal{F}$ in the decoding layers of $\boldsymbol{\epsilon_\theta}$ is expressed as:
\begin{equation}
    \mathcal{F} = \mathrm{F}_n(z_0; \theta) + \mathrm{Z}(\mathrm{F}_n(\mathcal{G}; \theta_{c}); \theta_{z}),
\end{equation}
\noindent where  $\mathrm{F}_n(\cdot; \theta)$ is a trained neural network, $\mathrm{Z}(\cdot;\cdot)$ denotes zero convolution layers with both weights and bias initialized to zeros, $\theta_{c}$ represents the parameters within  ControlNet, $ \theta_{z}$ is the parameters of zero convolution layers. We feed the latent feature map and the pose-conditioned inputs to the pre-trained $\boldsymbol{\epsilon_\theta}$ network and extract the multi-scale feature maps $\mathcal{F}_i$ from the last layer of each output block in different resolutions.  Our experimental observations indicate that the extracted informative features represent more information about the structures in abstract sketches, enhancing the accuracy of the subsequent SMPL regression for sketch input.

{In addition, we empirically find that the cross-attention maps $A_i \in \mathbb{R}^{|\mathcal{G}| \times H_i \times W_i}$ from the decoding layers of the U-Net $\boldsymbol{\epsilon_\theta}$ can provide occlusion-aware cues that indicate invisible parts and help to focus on the 2D skeleton condition information within the sketch. Therefore, we concatenate the feature maps with the cross-attention maps to generate the hierarchical feature maps $\mathcal{F} \leftarrow \{[\mathcal{F}_i, A_i]\}$, which incorporate explicit and implicit diffusion priors and thereby further enhance the performance of SMPL mesh regression.}


\subsection{SMPL Mesh Regressor} \label{sec: regressor}
In the last stage, an SMPL mesh regressor is proposed to predict 3D poses from the previously extracted features $\mathcal{F}$.
Specifically, $p_{\phi_3}(\boldsymbol{y} \!\mid\! \mathcal{F})$ refers to the prediction head that generates parameters of the body model from the hierarchical feature maps $\mathcal{F}$. We first lift the pose-guided 2D feature $\mathcal{F}$ to the 3D feature $\mathcal{F}_{3D}$, then extend the 2D features by incorporating 3D joint feature sampling. To integrate and align 2D and 3D features, we utilize a fusion transformer ~\cite{li2023jotr} to regress SMPL parameters. Moreover, we employ a pre-trained VQVAE~\cite{van2017neural}, which is trained on a large-scale motion dataset AMASS~\cite{mahmood2019amass} with extensive SMPL pose parameters to provide adequate human pose priors and preserve the correspondence of the VQGAN framework, which can obtain discrete representations of human poses. During the regression, the decoder of the VQVAE is utilized to get the pose parameters $\Theta$, while the shape parameters $\beta$ and camera parameters $c$ are directly predicted using linear layers.

\subsection{Objective Function} \label{sec: loss}

Unlike human pose estimation from real-captured photos, recovering human pose from artificial sketches in the literature is even more difficult due to the distorted proportions, perspective, and foreshortening.  Specifically, sketches often depict characters with unrealistic body shapes or exaggerated body proportions. Therefore, standard optimization methods that depend solely on 2D joint positions can result in inaccurate or unnatural outcomes. Through observing the artwork of human character drawing, three key elements are identified as crucial for addressing these issues: joint angle, foreshortening, and self-contacts. Building on our proposed human drawing dataset with 3D pose annotations, we introduce new methods to address this.

\paragraph{Joint Angle}
Character bones often appear longer than their actual length because of imprecision in drawings or the use of artistic interpretation ~\cite{hogarth1970dynamic, johnston1981illusion, stanchfield2007gesture} in human drawings. Due to inconsistent representations of bone length, directly utilizing absolute joint positions becomes impractical. And art literature has consistently highlighted the importance of accurately describing joint angles. Therefore, we expect the 3D bone projections to align with the bones depicted in 2D, ensuring that the reconstructed 3D joint angles have corresponding projections to the depicted 2D joint angles.

Our dataset provides precise annotations for the 2D joints $x^{2D}$, 3D joints $x^{3D}$, and exact SMPL pose parameters $\Theta$ of the characters in each drawing. For each bone $i$ connecting joints $j_1$ and $j_2$, we  represent its 3D vector as $\boldsymbol{b}_i^{3D} = {x}_{j_2}^{3D} - {x}_{j_1}^{3D}$, and its orthographic projection onto the screen as $\boldsymbol{b}_i^{2D}$. The 2D joints predicted by our algorithm are $\bar{x}^{2D}$, so the predicted vector corresponding to bone $i$ is $\boldsymbol {\bar{b}_i^{2D}} = \bar{x}_{j_2}^{2D} - \bar{x}_{j_1}^{2D}$,  and $\boldsymbol n$ represents the normal to the predicted bone $\boldsymbol {\bar{b}_i^{2D}}$. Guided by our principle of joint angle, the loss of parallelism between the projected 2D bones can be expressed as:
\begin{equation}
    \mathcal{L}_{\mathrm{parallel}} = \sum_i \left( \frac{\boldsymbol{b}_i^{2D}}{\|\boldsymbol{b}_i^{2D}\|} \cdot\boldsymbol{n} \right)^2,
     \label{eq:parallel}
\end{equation}
This skeleton parallelism loss enables a more reasonable and natural alignment of human joints in sketches than joint position loss.

\renewcommand{\arraystretch}{1.1}
\begin{table*}[t]
\centering
\caption{Quantitative comparison on the artist-designed dataset and the SKEP-120K validation set.(Unit: mm)}
\resizebox{0.95\linewidth}{!}{
\begin{tabular}{lccccccccc}
\toprule
\multirow{2}{*}{Method} & \multicolumn{3}{c}{Expert1} & \multicolumn{3}{c}{Expert2} & \multicolumn{3}{c}{SKEP-120K} \\ \cline{2-4 }  \cline{5-7} \cline{8-10}
 & MPVE\(\downarrow\) & MPJPE\(\downarrow\) & PA-MPJPE\(\downarrow\) & MPVE\(\downarrow\) & MPJPE\(\downarrow\) & PA-MPJPE\(\downarrow\) & MPVE\(\downarrow\) & MPJPE\(\downarrow\) & PA-MPJPE\(\downarrow\)  \\
\hline
PyMAF       \cite{zhang2021pymaf}      & 312.7  & 299.4   & 187.5   & 301.5 & 291.2   & 187.0  & 143.1    &  117.4  &  101.3 \\ 
EFT         \cite{joo2020eft}          & 144.6  & 144.4   & 98.9    & 168.8 & 158.9   & 103.4   &     158.7    &  133.1   & 111.6   \\
HybrIK      \cite{li2021hybrik}        & 348.7  & 345.5   & 199.6   & 365.2 & 352.0   & 208.5   &     211.0    &  177.7  &  144.0 \\ 
CLIFF       \cite{li2022cliff}         & 186.4  & 181.5   & 142.2   & 217.3 & 201.3   & 144.5   &     137.3    &  113.3  &  98.0  \\ 
HMR2.0      \cite{goel2023humans}      & 118.3  & 105.0   & 85.1    & 181.4 & 151.4   & 107.4  &     128.0    &  104.6  &  88.1  \\ 
MotionBERT  \cite{zhu2023motionbert}   & 170.6  & 165.2   & 120.1   & 189.1 & 172.2   & 127.6  &     124.4    &  99.4   &  83.6  \\ 
{DPMesh      \cite{zhu2024dpmesh}}       & {130.6}  & {121.5}   & {95.3}    &{169.9}  & {152.1}   & {103.2}  &     {123.4}    &  {98.0}   &  {81.1}  \\
{DPMesh(Retrained)      \cite{zhu2024dpmesh}}       & {127.7}  & {121.4}   & {94.1}    &{166.4}  & {147.1}   & {94.3}  &     {122.6}    &  {97.3}   &  {80.6}  \\
Sketch2Pose \cite{brodt2022sketch2pose}& 103.8  & 101.4    & 78.1   &\textbf{145.5}  & 135.9  & 86.8  &     152.1    &  125.9   &  100.3    \\  
Ours            & \textbf{103.1}& \textbf{95.7}  & \textbf{77.4} &146.5 & \textbf{131.5}  & \textbf{84.3} & \textbf{106.7}  & \textbf{87.7} & \textbf{72.6} \\
\bottomrule
\end{tabular}}
\label{tab:mannualdata-eval}
\end{table*} 

\paragraph{Foreshortening}

Empirically, artists typically do not rely on exact mathematical measurements for orthographic or perspective projections when creating drawings~\cite{hogarth1970dynamic, stanchfield2007gesture}. Thus, directly reconstructing 3D poses from predicted 2D poses frequently results in highly inaccurate estimations of the angles formed between the bones and the screen. 
For bone $i$, the angle between the character in the drawing and the screen can be represented as the angle between the 3D vector of the skeleton and the 2D vector of its projection. The foreshortening loss for the skeleton can thus be formulated as:
\begin{equation}
    \mathcal{L}_{\mathrm{f}} = \sum_i \left( \frac{\| \boldsymbol{b}_i^{3D} \|}{\|\boldsymbol{b}_i^{2D}\|} - \frac{\| \boldsymbol{\bar{b}_i^{3D}} \|}{\|{\boldsymbol {\bar{b}_i^{2D} \|}}} \right)^2,
    \label{eq:foreshortening}
\end{equation} 
where $\mathcal{L}_{\mathrm{f}}$ is the skeleton-screen angle's cosine.


\paragraph{Self-contacts}
Self-contacts are prevalent in common human poses, which are revealed by prior works~\cite{hogarth1970dynamic}. We hypothesize that human observers often rely on perceived self-contacts to solve the problem of depth ambiguity and link touching body parts to similar depths. Previous works focus on optimizing regions based on manually annotated self-contact areas. They enforce physical contact between pairs of vertices by mapping each contact region onto the vertices of the roughly aligned SMPL mesh. In contrast, our dataset includes accurate SMPL pose parameters for the human body in sketches, which provides correct relative depth and joint positions of the character skeleton. In our method, we replace the previous self-contact loss with the SMPL pose parameter loss, which is calculated by $L_1$ loss between the predicted SMPL pose parameter and the ground-truth SMPL pose parameter, thus supervising the human mesh to recover the correct positions.

Instead of the previous position-based reprojection loss, our overall training objective in our method is defined as:
\begin{equation}
\mathcal{L} =\lambda _1\mathcal{L} _{\mathrm{parallel}}+\lambda _2\mathcal{L} _{\mathrm{f}}+\lambda _3\mathcal{L} _{\mathrm{pose}} + +\lambda _4\mathcal{L} _{\mathrm{shape}},\label{eq:loss}
\end{equation} \label{eq:9}
where $\mathcal{L} _{\mathrm{pose}}$ is the SMPL pose parameter loss, and $\mathcal{L} _{\mathrm{shape}}$ is the SMPL shape parameter loss, $\lambda_1$, $\lambda_2$, $\lambda_3$, $\lambda_4$ are set to 3, 3, 2, 1.

\section{Experiments}
\label{sec:exp}

\subsection{Implementation Details}
\paragraph{Datasets}
We utilize two datasets for performance evaluation: 

\textbullet \quad The artist-designed dataset is provided by Sketch2Pose~\cite{brodt2022sketch2pose}, which contains six sketches with corresponding 3D poses manually modeled by two artists that best align with the artist's intentions.  The merit of this validation set lies in its accurate representation of the ideal 3D pose intended by the artists, whereas its limitation is the scant data volume, comprising merely six very challenging poses.  

\textbullet \quad The \datasetname validation set is created using the method outlined in Section~\ref{sec: dataset}. We invite experienced 3D modelers to manually sieve through and eliminate inaccurate data to guarantee high quality. This validation set contains 600 validation tuples, with 100 tuples for each of the six styles.
Owing to its comprehensive coverage of various sketch styles and extensive data volume, it is well-suited for assessing the generalization capability.
\paragraph{Metrics}
We adopt three standard metrics for 3D pose estimation: Mean Per Joint Position Error (MPJPE) and Procrustes-Aligned Mean Per Joint Position Error (PA-MPJPE) to evaluate the accuracy of the predicted 3D joint positions, and Mean Per Vertex Error (MPVE) to measure the accuracy of 3D mesh reconstruction in sketches. Quantitative metrics are only part of the evaluation for the prediction of 3D poses in sketches. We place greater emphasis on whether the visualized results align more closely with the artist’s original intent. 
\paragraph{Training Details}
We train three models for data creation and sketch-to-pose prediction: ControlNet to synthesize sketch data, ViTPose for 2D keypoint detection, and our core sketch-to-pose network. For ControlNet, condition maps and sketches are padded to $512\times512$. Then, BLIP2~\cite{li2023blip} is leveraged to generate prompt labels for the sketches. 
ViTPose is fine-tuned separately on our collected and synthesized sketch data and remains frozen during core-model training, serving solely as a 2D joints extractor.
The core model training process consists of two stages. All training and experiments run on 4 Nvidia A6000 GPUs. The supplementary material provides more training details.

\begin{table}[t]
\caption{Quantitative comparison of ablation study.}
\resizebox{0.85\linewidth}{!}{
\begin{tabular}{@{}lccc@{}}
\toprule
Method vs Expert 1                   &  MPVE↓    &  MPJPE↓  & PA-MPJPE↓ \\ \midrule
\textit{w\( /\)o} $L_{\mathrm{parallel}}$   &  169.8    &  165.7   &  102.5      \\
\textit{w\( /\)o} $L_{\mathrm{f}}$          &  117.7    &  110.2   &  84.4     \\
\textit{w\( /\)o} $L_{\mathrm{pose}}$       &  121.4    &  117.6   &  86.7    \\ 
\textit{w\( /\)o} $L_{\mathrm{parallel}, \mathrm{f}, \mathrm{pose}}$   & 113.7     &  107.3   &   85.6     \\ 
\textit{w\( /\)o} $A_i$   & {104.9}     &  {97.1}  &   {79.6}     \\
\textit{w\( /\)o} $J^{2D}$   & {117.4}     &  {113.9}   &   {89.6}     \\
\textit{w\( /\)o} Data Curation   & {104.8}     &  {99.0}  &   {82.7}     \\
Ours (Full)             & \textbf{103.1} & \textbf{95.7} & \textbf{77.4}  \\ \bottomrule
\end{tabular}}
\label{tab:ablation}
\end{table}

\begin{table}[t]
\caption{Runtime of our method.}
\resizebox{0.95\linewidth}{!}{
\begin{tabular}{@{}lcccc@{}}
\toprule
Method     & I (Sec. 4.1) & II (Sec. 4.2) & III (Sec. 4.3) & Total Time \\ \midrule
{PyMAF$^*$}  & {0.09s}  & {-} & {0.01s}  & {0.10s} \\
{EFT$^*$}  & {0.34s}  & {-} & {3.23s}  & {3.57s} \\
{HyBrIK$^*$}  & {0.03s}  & {0.06s} & {0.08s}  & {0.17s} \\
{CLIFF$^*$}  & {2.60s}  & {1.21s} & {0.04s}  & {3.85s} \\
{HMR2.0$^*$}  & {0.52s}  & {-} & {0.03s}  & {0.55s} \\
{MotionBERT$^*$}  & {0.04s}  & {-} & {0.15s}  & {0.19s} \\
{DPMesh$^*$}  & {0.05s}  & {0.06s} & {0.03s}  & {0.14s} \\

Sketch2Pose  & 4.75s         & 32.98s         & 30.15s          & 67.57s     \\ 
Ours   & 0.04s         & 0.05s         & 0.03s          & 0.12s      \\ \bottomrule
\end{tabular}}
\\$^*$ means the model is not for sketches but for regular photos.
\label{tab:runtime}
\end{table}

\subsection{Quantitative Comparison} \label{sec:Quantitative}
Quantitative comparisons on the artist-designed dataset and \datasetname validation set are shown in Tab.~\ref{tab:mannualdata-eval}.  
The artist-designed dataset contains six real-world sketches with expert-annotated 3D poses, all of which are challenging non-daily poses. On this dataset, our model achieves the best overall performance of all metrics.  Although gains on charcoal sketches are modest relative to Sketch2Pose and we even perform slightly worse on one expert-based metric, we reach this accuracy in roughly $1/500$ of Sketch2Pose’s runtime. 


The \datasetname validation set contains sketches generated outside of our training set. It includes six sketch styles, each containing 100 images, all following a real-world sketch distribution. Across all styles, our method significantly outperforms prior approaches and achieves the best overall results.


These results show that our method balances high accuracy with strong efficiency. Its generalization surpasses both the generic pose estimation algorithm and Sketch2Pose algorithm. To ensure a fair evaluation, all methods use the same 2D inputs produced by our trained ViTPose model. For generic pose estimation, we select the current leading method DPMesh and retrain it with our data. The performance gains verify the effectiveness of our dataset. Moreover, with identical training data, our method still outperforms alternative approaches. Furthermore, its fast inference enables efficient application to video, beyond static images.

\subsection{Qualitative Comparison} \label{sec:Qualitative}

\begin{figure*}[]
    \centering
    \includegraphics[width=0.9\textwidth]{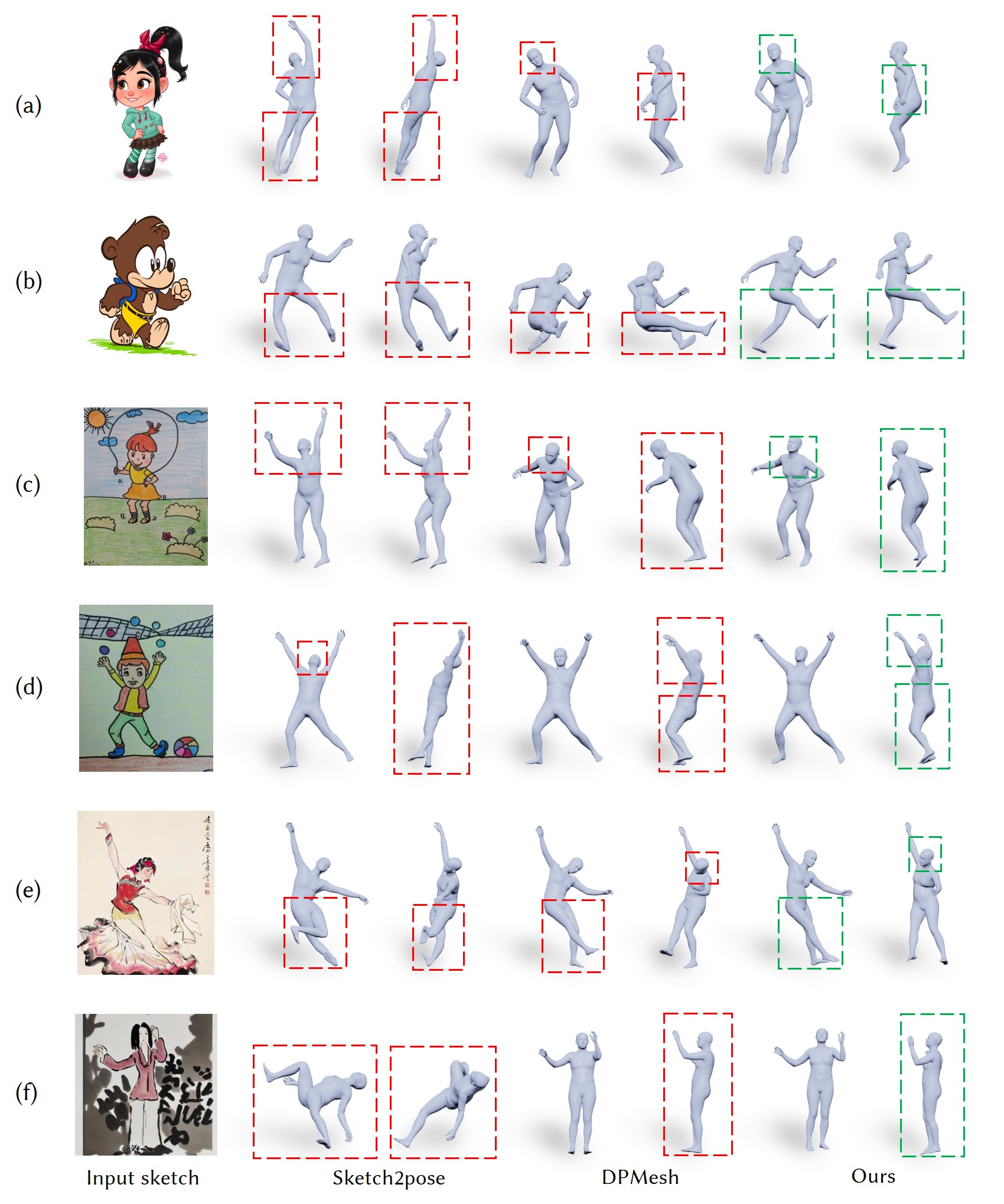}
    \caption{\textbf{Qualitative Comparison of Multiple Sketch Styles.}  Our proposed model accurately predicts real human body proportions in cartoon images and outperforms other methods in various sketch styles. Its high performance across multiple sketch styles is attributed to our three-stage pose prediction network design and diverse dataset with perturbations.  The red dashed box highlights the unreasonable 3D human pose estimation.
    }
\label{fig: result_style}
\end{figure*}

We have visualized comparisons across diverse sketch styles in Fig.~\ref{fig: result_style}. Notably, (a) - (b) demonstrate that our model can predict poses consistent with real human body proportions even for cartoons with exaggerated proportions. (c) - (f) highlight that our method yields more accurate predictions for cartoons, children's drawings, stick figures, ink paintings, and oil paintings. Sketch2Pose struggles beyond charcoal sketches, and the generic image-to-pose baseline DPMesh degrades markedly when dealing with various styles. By contrast, only our method sustains high performance across multiple sketch styles. We attribute this to our three-stage pose prediction network design for sketch feature extraction and synthesized dataset with perturbations.

\begin{figure*}[]
    \centering
    \includegraphics[width=1.0\textwidth]{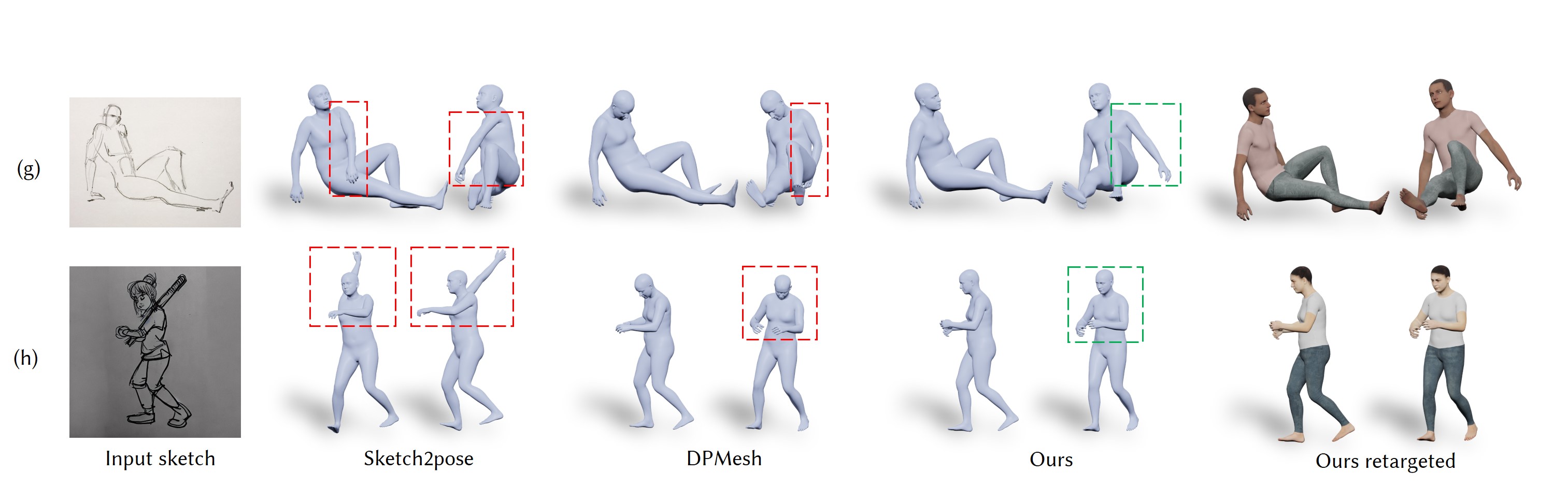}
    \caption{\textbf{Qualitative Comparison of Challenging Poses.}  Our method successfully recovers plausible poses even when character parts are obscured or lines overlap, as illustrated in specific scenarios. These results demonstrate that our approach better meets artists' needs and predicts 3D human poses with higher accuracy.  The red dashed box highlights the unreasonable 3D human pose estimation. And our results can be seamlessly applied to a custom character using standard tools.
    }
\label{fig: result_pose}
\end{figure*}

Fig.~\ref{fig: result_pose} presents the predictions for challenging poses. Specifically, (g) illustrates a scenario where one of the character's arms is fully occluded, yet our method infers a plausible pose. In (h), overlapping character lines with a handheld object simplify structure, but the pose is still correctly predicted. These results indicate that our approach better meets artists' creative needs and predicts 3D human poses more accurately. We also retarget predicted poses to custom characters as shown in Fig.~\ref{fig:teaser}. Our method facilitates frame-by-frame prediction due to its efficient inference speed, thereby significantly improving the applicability and effectiveness of sketch-to-pose.

\begin{figure*}[]
    \centering
    \includegraphics[width=0.95\textwidth]{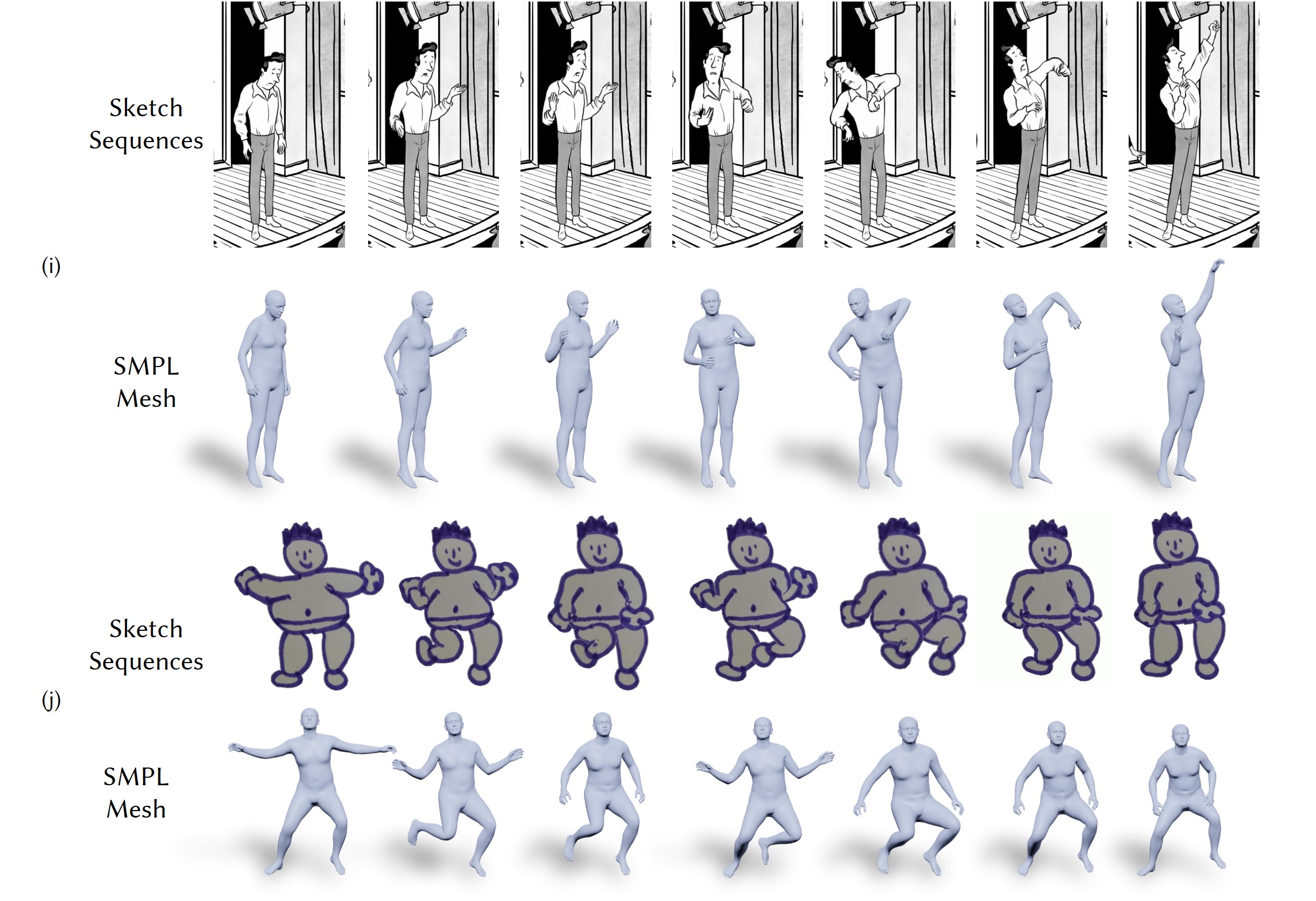}
    \caption{\textbf{Results on Sketch Video Input.}  The frame-by-frame prediction results show our method's generalization capability in extracting human poses from sketch videos.
    }
\label{fig: result_video}
\end{figure*}

Fig.~\ref{fig: result_video} presents the frame-by-frame results of our method applied to continuous line animations, demonstrating the generalization capability for pose estimation from videos. In addition, 
The supplementary material provides a user study for subjective evaluation. 


\subsection{Ablation Study} 

We perform an ablation study on the artist-designed dataset to evaluate the efficacy of our proposed loss terms. Specifically, we ablate the loss terms $\mathcal{L}_{\mathrm{parallel}}$, $\mathcal{L}_{\mathrm{f}}$, and $\mathcal{L}_{\mathrm{pose}}$ from Eq.~{\eqref{eq:9}}, and replace all proposed loss terms with the joint-distance-based loss used in DPMesh, retraining the model with all other settings the same. The evaluation results reported in Tab.~\ref{tab:ablation} reveal a performance degradation across all ablation settings, indicating the indispensable contribution of each ablated loss term to the overall model performance.

Next, we perform further ablations on the network design. Specifically, we remove the 2D joint prediction $J^{2D}$ from the 2D guidance extractor and exclude the cross-attention maps $A_i$ from the sketch feature extractor. Tab.~\ref{tab:ablation} shows a decline in model performance, demonstrating the effectiveness of our network design.
Moreover, the quality of generated sketches is important for model performance. We fine-tune ControlNet on real sketches to synthesize high-quality training data, and subsequently apply manual curation to further enhance data quality. To validate the effect of sketch quality, we retrain our model on the pre-curation set and find that training on the curated set yields superior performance, showing the importance of generated-sketch quality for overall effectiveness.
\subsection{Runtime Analysis} 
\label{sec:runtime}

The step-by-step runtimes of our method and Sketch2Pose are reported in Table~\ref{tab:runtime}. Sketch2Pose involves a three-stage process that differs from our design, including: (I) a 2D pose and initial 3D mesh estimator; (II) an optimizer to correct errors in the estimated 3D pose; and (III) a final refinement to better align the result with the stylistic and structural constraints of the sketch.  Since Sketch2Pose leverages iterative optimization approaches to solve all three stages, it incurs substantial computational cost.
By contrast, our approach achieves an over $500\times$ speedup while delivering superior performance compared to Sketch2Pose. This improvement is primarily attributed to our efficient feed-forward neural network design. In addition to the comparison with Sketch2Pose, we also compare our method with other approaches used for human pose estimation from regular photos. The results demonstrate that our method achieves a significant advantage in runtime, matching or even surpassing the methods of pose estimation from regular photos.

\begin{figure}[!h]
    \centering
    \includegraphics[width=1.0\linewidth]{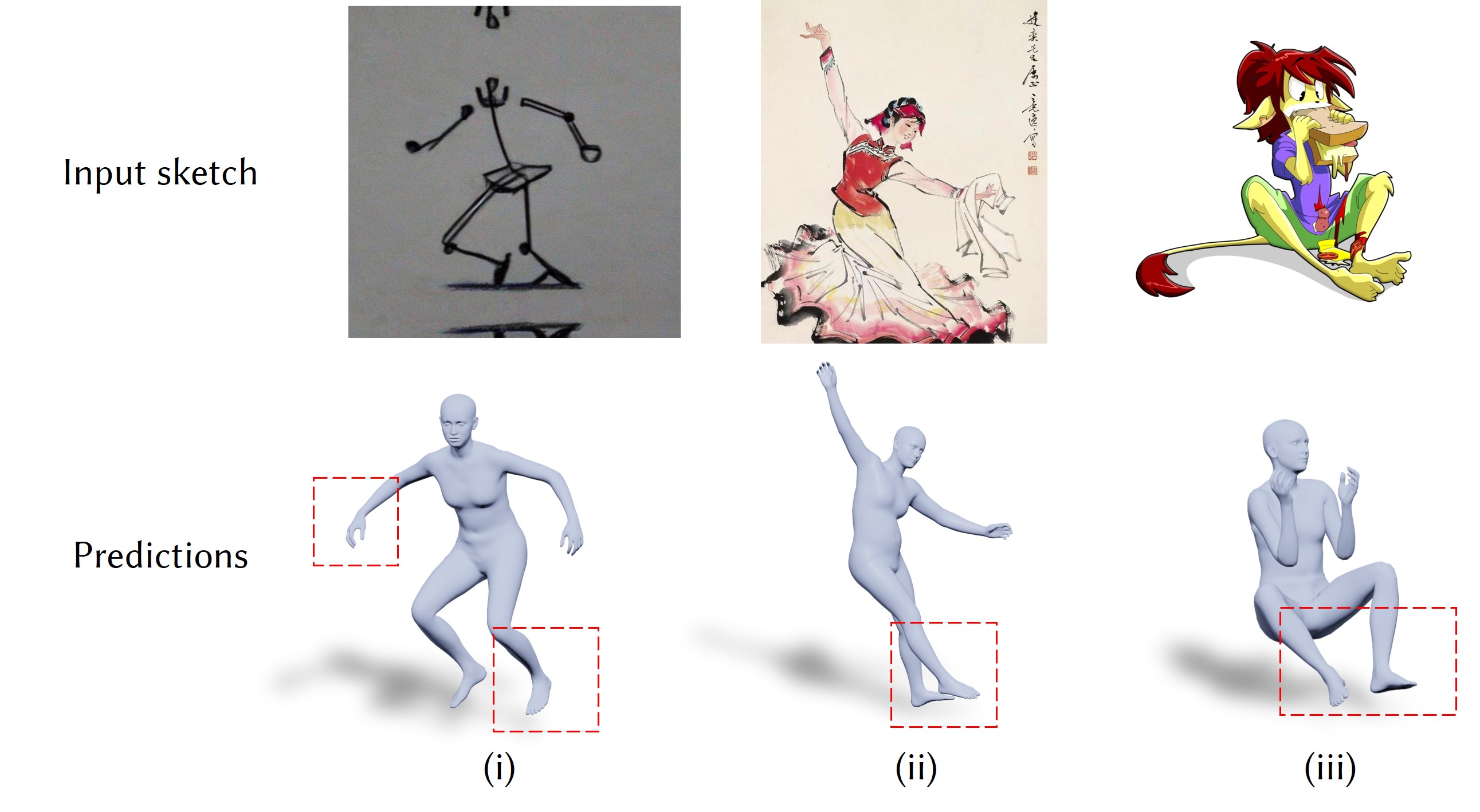}
    \caption{Failure cases. Our model may predict inaccurate terminal joints.
    }
\label{fig:limitations}
\end{figure}

\section{Conclusion} 
\label{sec:con}

We present a novel approach to estimating human poses from sketches. By adopting a learning-by-synthesizing strategy, we have synthesized a large-scale, customized sketch-pose dataset tailored for this task, significantly enhancing our model's generalization capabilities across various sketch styles.
Furthermore, the proposed feed-forward structured network has markedly improved the speed of sketch-to-pose estimation.
\paragraph{Limitations.} 

Terminal joints (e.g., those in the hands and feet) in sketch-based human pose estimation remain particularly challenging to predict, owing to multiple factors. Firstly, sketches are inherently abstract and often lack fine structural detail. For instance, a single limb stroke may ambiguously depict either the contour of a forearm or an extended hand as shown in Fig. ~\ref{fig:limitations} (i), providing only limited discriminative cues. 
Secondly, terminal joints in sketches are often occluded or have ambiguous depth cues. While our method leverages a pretrained diffusion model to infer plausible positions for occluded terminal joints, its predictions remain unnatural as shown in Fig.~\ref{fig:limitations} (ii).
Thirdly, the hierarchical structure of the human skeleton means that minor errors at proximal joints can propagate along the kinematic chain, leading to disproportionately large errors at distal joints as shown in Fig. ~\ref{fig:limitations} (iii).
Collectively, these factors result in lower estimation accuracy for terminal joints.

Our current work focuses on sketch-driven 3D human pose estimation. Subject-specific body-shape reconstruction is out of the scope of our research. We have conducted further analysis of this limitation in the supplementary materials, and addressing this issue has become a priority for future research.

\begin{acks}
This study was funded by NKRDC 2022YFF0902200 and Jiangsu Broadcasting Corporation.
\end{acks}



\newpage
\bibliographystyle{ACM-Reference-Format}
\bibliography{main}

\newpage

\end{document}